# RAG for Effective Supply Chain Security Questionnaire Automation

*Completed Research Paper*

Zaynab Batool Reza[1*], Abdul Rafay Syed[1], Omer Iqbal[1], Ethel Mensah[1], Qian Liu[1], Maxx Richard Rahman[1,2], Wolfgang Maass[1,2]

[1] Saarland University, Germany
[2] German Research Center for Artificial Intelligence (DFKI)
[*] Corresponding Author: zare00001@stud.uni-saarland.de

## Introduction

In the global economy, ensuring supply chain integrity is crucial for businesses across industries, to protect against disruptions, cyber threats, and data breaches (Saenz et al. 2015). Companies face challenges in maintaining security standards, particularly in managing varied and complex questionnaires from partners and regulators, which are time-consuming and repetitive for IT security teams.

The primary issue addressed in this paper is enhancing consistency in responses to supply chain security questionnaires. These repetitive tasks drain resources and risk security compromises due to potential errors or omissions (Saenz et al. 2015). We propose a solution that automates the interpretation of these varied questionnaires, providing accurate, consistent answers based on predefined security protocols.

Our research focuses on developing an application capable of parsing different document formats, understanding questionnaire contexts, and generating precise responses. We introduce a robust NLP model to recognize and interpret security-related inquiries, integrate this model into a user-friendly internal tool, and implement analytics to assess and refine response accuracy. Additionally, we develop automated testing frameworks to ensure the application's reliability and consistency over time.

## Related Works

The development of automated form-filling systems has gained significant attention due to their potential to enhance efficiency and accuracy across industries. This section reviews recent advances, focusing on large language models (LLMs) and retrieval-augmented generation (RAG) for automated questionnaire completion based on internal policies.

Several studies have explored LLMs and RAG in different domains. Fuchs et al. (2021) introduced Form-BERT, a transformer-based model for auto-completing e-commerce forms by predicting attribute values, showcasing deep learning's specialized role in form automation. In contrast, Bucur (2023) integrated a knowledge base with an LLM for broader form completion tasks, emphasizing external knowledge sources to enhance accuracy, unlike Form-BERT's more focused use case.

Lewis et al. (2020) and Jeong (2023) examined integrating retrieval systems with LLMs to enhance response generation in knowledge-intensive tasks. Lewis et al. highlighted external data's role in informing LLM responses for handling diverse questionnaire formats, while Jeong focused on LangChain's use of internal business data to improve response accuracy. Both approaches demonstrate the importance of combining external knowledge with LLMs for better performance in complex tasks.

Johnson et al. (2017) provided the FAISS library for efficient similarity searches in large datasets of embeddings, supporting the rapid retrieval of relevant policy information—critical for improving the accuracy and efficiency of large-scale data retrieval systems like ours.

Recent works such as Atlas by Izacard et al. (2022) and a survey by Fan et al. (2024) emphasized the significance of RA-LLMs. Atlas showed retrieval-augmented models' effectiveness in few-shot learning, especially for knowledge-intensive tasks, without extensive parameters. Fan et al. offered a comprehensive overview of RA-LLMs, discussing their architectures, training strategies, and application areas, highlighting how integrating external knowledge helps overcome LLM limitations like hallucinations and outdated data.

These studies emphasize the versatility of RAG with LLMs for handling complex and varied questionnaire formats. Our project builds on these methodologies, focusing on internal policies as the primary data source to streamline form-filling within organizations.

## Model Development

### *Problem Statement*

The field of supply chain security requires extensive documentation and strict adherence to internal policies. IT security teams face the challenge of completing complex, time-consuming questionnaires that assess compliance with security policies and potential risks. These questionnaires, prone to human error and quality inconsistencies, are detailed and demanding. Our model leverages document embeddings and queries (q) stored in an Elasticsearch Vector Database (Elastic, 2023), guiding the retrieval function of an LLM model, which fetches the most relevant documents (D) for each query. The generative model then creates responses based on the retrieved documents.

## System Architecture and Workflow

Our model, QuestSecure, automates questionnaire completion using a blend of algorithms and formulas to ensure accuracy and efficiency. Key components include the Input Module, which preprocesses documents and converts text into vector representations via LangChain tools (LangChain, 2023). These vectors are stored in an Elasticsearch-managed database for fast retrieval. The Retrieval Module identifies relevant documents, and the Response Generation Module, powered by a Retrieval-Augmented Generation (RAG) framework, generates accurate, natural responses based on retrieved content and the query.

LangChain optimizes the workflow, from embedding generation to response generation, ensuring a streamlined process. This architecture maintains data integrity and relevance while facilitating smooth communication between modules.

Interaction is enhanced through a Streamlit-based frontend (Streamlit, 2023), improving user experience and scalability. The system supports load balancing and distributed computing to handle growing data needs. Figure 1 outlines the system architecture, demonstrating how components work together to efficiently process queries and generate responses, improving IT security teams' efficiency while maintaining data privacy standards.

$$Embedding = f(CleanedText)$$

$$D_{relevant} = Retrieve(q, E)$$

$$Response = Generate(D_{relevant}, q)$$

Here, $Embedding$ refers to the mathematical representation of text as dense vectors that capture meaning. $D_{relevant}$ represents the most relevant documents retrieved from the database based on the query (q) and embeddings (E), using the function $Retrieve(q, E)$. The $Response$ is the final output, generated by $Generate(D_{relevant}, q)$, which combines the retrieved documents with the query to produce a contextually accurate answer.

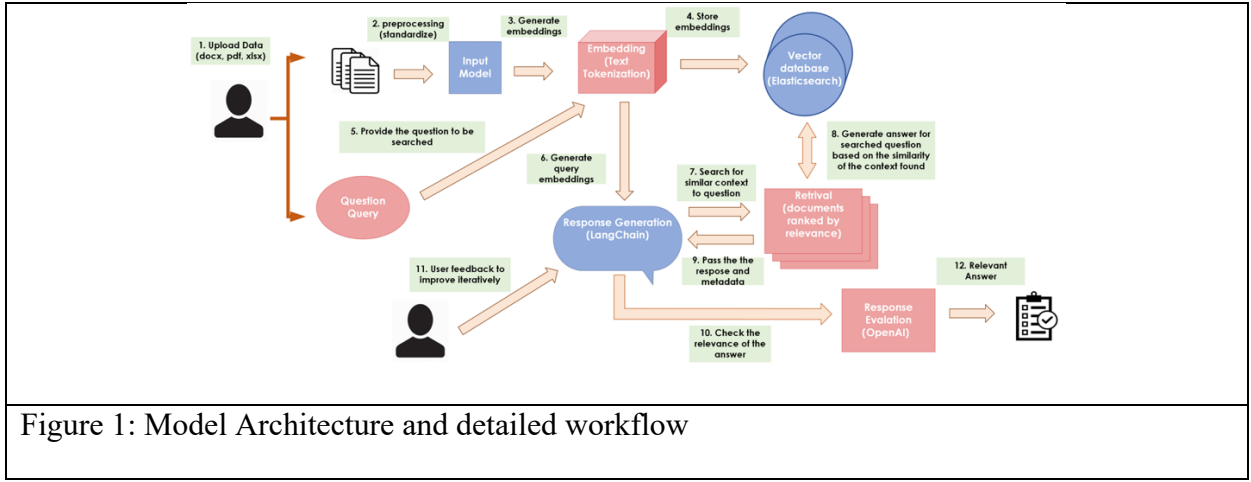

Figure 1: Model Architecture and detailed workflow

## Experiments

### Data

We use a dataset from the German Federal Office for Information Security (BSI) (Bundesamt für Sicherheit in der Informationstechnik - BSI), drawn from the IT-Grundschutz Compendium. This collection includes German-language PDFs and Excel files with security protocols, guidelines, and questionnaires designed for IT security in various organizations. The dataset's complexity, including tables, text, and images, is ideal for training our NLP model to handle security-related queries. Through preprocessing, the model extracts semantic information, enhancing its ability to respond effectively to security questionnaires

### Preprocessing

In preprocessing, we use LangChain tools to process PDFs, Excel, images, and CSVs. Text is extracted with specialized modules, cleaned, and normalized for consistency. The cleaned text is then processed by an LLM to create semantic embeddings, capturing essential details for accurate data interpretation. These embeddings are stored in Elasticsearch for quick retrieval in our Retrieval-Augmented Generation framework, laying the groundwork for efficient responses to security questionnaire.

### *Baseline Models*

To establish a robust baseline for our experiments, we employ PyPDFLoader (LangChain, 2023a) and CSVLoader (LangChain, 2023b) to accurately load PDF and CSV documents, ensuring they are correctly parsed and formatted for further processing. For document segmentation, we use RecursiveCharacterTextSplitter (LangChain, 2023c) and CharacterTextSplitter (LangChain, 2023d), setting each chunk to 150 characters with zero overlap for comprehensive document coverage.

In the embedding generation phase, we use the MiniLLM model from Nomic AI (2024), known for its efficient production of high-quality, semantically rich embeddings. For retrieval, we implemented a basic similarity search with k=20 to identify the top 20 documents matching the query's context.

For response generation, we utilize the LlaMa 3 model, adept at generating instruction-based responses. This process, streamlined by a standard Retrieval-Augmented Generation (RAG) prompt from LangChain's hub, allows us to create structured queries for accurate and relevant response generation.

### *Post Processing*

In the post-processing phase, we refine model responses to ensure clarity and relevance. We remove repetitive elements and redundant phrases that do not contribute to a concise answer. Additionally, we filter out segments that stray from the question's language and context, maintaining focus and accuracy. This polishing ensures clear, contextually appropriate, and applicable responses, enhancing the system's effectiveness in handling supply chain security questionnaires.

### *Experimental Setup*

We experiment with chunk sizes from 150 to 512 characters and overlaps up to 100 to ensure the right balance between text granularity and contextual coherence. Different retrieval mechanisms, including similarity search and Maximal Marginal Relevance (MMR) (LangChain, 2023e), are tested to source the most contextually appropriate information. We vary the number of documents retrieved to refine our process further.

Prompt engineering is key; we modify the location and details within the prompts to test how these changes affect the clarity and directness of responses. We experiment with different configurations and language-specific instructions to enhance output accuracy.

Additionally, we use different models like LlaMa and Mistral Instruct to handle complex queries and generate context-aware answers.

These experiments refine our approach and enhance our ability to automate responses to security questionnaires, establishing a robust framework that adapts to diverse document types and processing needs, thus improving our system's efficacy.

### *Evaluation Metrics*

We evaluate our proposed RAG-based approach using both semantic-based automatic metrics and human-like evaluation via an LLM.

**Semantic-Based Automatic Metrics**: We employ BertScore (Zhang et al., 2020) and METEOR (Banerjee & Lavie, 2005) for their effectiveness in capturing the semantic richness of responses. BertScore utilizes BERT's contextual embeddings to provide precision, recall, and F1 scores, assessing text similarity between generated and reference responses. METEOR enhances n-gram overlap methods by considering alignment and linguistic nuances like synonyms and paraphrasing, offering a comprehensive measure of linguistic richness.

**G-Eval with EM German Mistral Model**: Complementing automatic metrics, G-Eval (Liu et al., 2023) applies a powerful LLM, the EM German Mistral Model, to score responses on Context Precision, Context Recall, Faithfulness, and Answer Relevancy. These metrics evaluate the accuracy, coverage, truthfulness, and relevance of the responses, ensuring they align closely with the contextual demands of the queries.

By integrating G-Eval, we incorporate a human-like perspective into our evaluation, addressing nuances potentially overlooked by automated metrics. This dual approach provides a thorough assessment of the RAG model, ensuring responses are not only accurate but contextually apt and linguistically coherent.

## Results

We evaluate the performance of our proposed approach and all the baseline models. Figure 2 shows the results of different model configurations, and the percentage of valid responses returned. We use these results to determine which prompting technique is the most suitable. The configurations can be interpreted as given in Table 1.

| Configuration | Retrieval Technique | LLM Model | Prompting Technique | Chunk Size | Excel Handling |
|---|---|---|---|---|---|
| SLOBE | Similarity | LLAMA | Start of Prompt (O) | 150 | Excel Separate |

| SLOB | Similarity | LLAMA | Start of Prompt (O) | 150 | Standard |
|------|-----------|-------|---------------------|-----|----------|
| SLNC | Similarity | LLAMA | Start and End (N) | 512 | Standard |
| SMNC | Similarity | Mistral | Start and End (N) | 512 | Standard |
| MLNC | Maximal Marginal Relevance (MMR) | LLAMA | Start and End (N) | 512 | Standard |
| MMNC | Maximal Marginal Relevance (MMR) | Mistral | Start and End (N) | 512 | Standard |
| SLOC | Similarity | LLAMA | Start of Prompt (O) | 512 | Standard |
| SMOC | Similarity | Mistral | Start of Prompt (O) | 512 | Standard |
| MLOC | Maximal Marginal Relevance (MMR) | LLAMA | Start of Prompt (O) | 512 | Standard |
| MMOC | Maximal Marginal Relevance (MMR) | Mistral | Start of Prompt (O) | 512 | Standard |

Table 1: Different configurations used for experimentation.

Figure 2 shows the percentage of valid responses across different model configurations. Each bar represents a specific combination of retrieval technique, LLM model, prompting technique, and chunk size.

Configurations using prompt O (start of the prompt) consistently have higher valid response rates compared to prompt N (start and end). The highest rate is for SLOC, which uses Similarity retrieval, LLAMA, and prompt O with a 512-token chunk size. Based on these results, we focus further testing on configurations with prompt O to streamline evaluation and prioritize configurations with higher valid response rates.

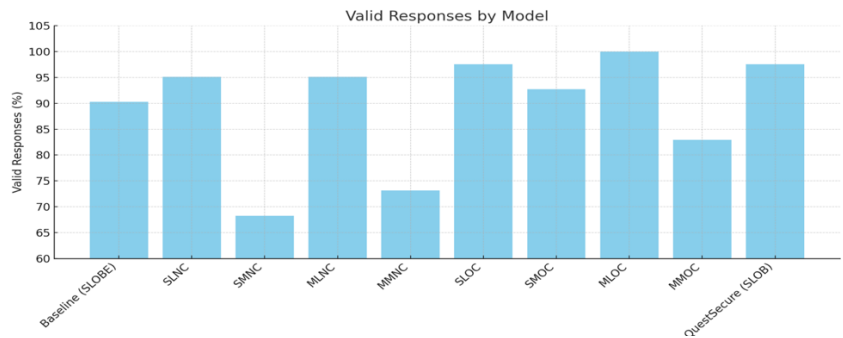

Figure 2: Valid Responses

The data in Table 2 shows QuestSecure's superiority in generating responses, with improvements in METEOR and BertScore metrics. This reflects the model's enhanced ability to understand and generate semantically rich and precise responses.

| Metric/ Model | METEOR | BertScore (Precision) | BertScore (Recall) | BertScore (F1) |
|---|---|---|---|---|
| Baseline (SLOBE) | 0.114 | 0.634 | 0.655 | 0.643 |
| QuestSecure (SLOB) | 0.139 | 0.693 | 0.692 | 0.692 |
| SLNC | 0.128 | 0.694 | 0.689 | 0.691 |
| SMNC | 0.093 | 0.699 | 0.675 | 0.686 |
| MLNC | 0.133 | 0.694 | 0.689 | 0.691 |
| MMNC | 0.111 | 0.698 | 0.682 | 0.689 |
| SLOC | 0.138 | 0.683 | 0.691 | 0.687 |
| SMOC | 0.101 | 0.683 | 0.6708 | 0.676 |
| MLOC | 0.149 | 0.688 | 0.694 | 0.691 |
| MMOC | 0.108 | 0.681 | 0.6702 | 0.674 |

Table 2: Comparison of QuestSecure with other configurations for Base Metrics

Table 3 provides insight into the human-like quality of the responses, evaluating them on criteria that mimic human judgment. The QuestSecure configuration notably excels in all evaluated aspects, particularly in Contextual Precision and Answer Relevancy, affirming its applicability and effectiveness in real-world settings. Due to limited resources, we were only able to run this on 3 configurations.

| Performance Metric | QuestSecure(SLOB) | SLNC | SLOC |
|---|---|---|---|
| G-Eval (Contextual Precision) | 4.0535 | 3.568 | 3.6206 |
| G-Eval (Contextual Recall) | 3.689 | 3.281 | 3.339 |
| G-Eval (Faithfulness) | 3.5 | 2.35 | 2.078 |
| G-Eval (Answer Relevancy) | 4.5 | 3 | 3.68 |

Table 3: G-Eval Results

## Discussion

Our results reveal key insights into the effectiveness of different configurations for automating supply chain security questionnaires. Prompts with instructions only at the start (prompt O) consistently led to higher valid response rates, suggesting that simplifying instruction placement enhances accuracy by reducing cognitive load and irrelevant content generation.

Maximal Marginal Relevance (MMR) proved particularly effective in tasks requiring precise differentiation between similar documents, demonstrating its strength in high-precision contexts. Larger chunk sizes also improved performance by capturing more contextual information, leading to higher METEOR and BertScore (Recall) metrics due to enhanced semantic richness and better query alignment.

The Llama model outperformed Mistral Instruct, likely due to its extensive training on diverse datasets, allowing it to better handle the complexity of security-related queries.

G-Eval metrics provided a nuanced view of model performance, assessing contextual precision, recall, faithfulness, and relevancy of responses. While the basic configuration scored well, QuestSecure (SLOB) showed notable improvements across all G-Eval metrics. Further testing with more resources could provide deeper insights into other configurations.

## Conclusion

Our study demonstrates the effectiveness of Natural Language Processing (NLP) and Retrieval-Augmented Generation (RAG) frameworks in automating responses to supply chain security questionnaires, significantly enhancing efficiency and accuracy. Simplified prompts, advanced retrieval techniques like MMR, and larger data chunk sizes have collectively contributed to improved response quality. The Llama model has shown exceptional performance due to its comprehensive training on diverse datasets. This reduces the burden on IT security teams and minimizes error potential, thereby improving compliance and security standards. The inclusion of human-like evaluation metrics such as the G-Eval ensures that the automated responses align with nuanced human judgment, essential for maintaining response reliability in security settings.

***Limitations and Future Work***

Our study faced several limitations that affected the depth of our experimentation. Resource constraints limited our ability to test a wider range of model configurations, potentially missing insights into optimal settings. We also lacked access to more advanced LLMs, such as GPT-4 from Azure OpenAI, which restricted our ability to fully exploit the latest NLP advancements. Enhanced access to these models could significantly improve the accuracy and contextual relevance of the automated responses.

Looking ahead, future work could focus on integrating more advanced LLMs to further refine response quality, benefiting from the latest AI and machine learning advancements. With increased computational resources, we could also explore more thorough hyperparameter tuning, optimizing factors like chunk size and document retrieval through methods like random search.

## References

Banerjee, S., & Lavie, A. (2005). METEOR: An automatic metric for MT evaluation with improved correlation with human judgments. *Proceedings of the ACL Workshop on Intrinsic and Extrinsic Evaluation Measures for Machine Translation and/or Summarization*, 65-72. https://aclanthology.org/W05-0909.pdf

Bucur, M. (2023). *Exploring large language models and retrieval augmented generation for automated form filling*. University of Twente.

Bundesamt für Sicherheit in der Informationstechnik. (n.d.). IT-Grundschutz compendium.

Bundesamt für Sicherheit in der Informationstechnik. Retrieved from

https://www.bsi.bund.de/DE/Themen/Unternehmen-und-Organisationen/Standards-und-Zertifizierung/IT-Grundschutz/IT-Grundschutz-Kompendium/Hilfsmittel-und-Anwenderbeitraege/Recplast/recplast_node.html

Carbonell, J., & Goldstein, J. (1999). The use of MMR, diversity-based reranking for reordering documents and producing summaries. *Proceedings of the 22nd Annual International ACM SIGIR Conference on Research and Development in Information Retrieval*, 290-298.

https://doi.org/10.1145/290941.291025

Chase, H. (2022). LangChain [Computer software]. GitHub. https://github.com/langchain-ai/langchain

Elastic. (2023). *Elasticsearch: The official distributed search & analytics engine*. Elastic. https://www.elastic.co/elasticsearch

Fan, W., Ding, Y., Ning, L., Wang, S., Li, H., Yin, D., Chua, T. S., & Li, Q. (2024). A survey on RAG meeting LLMs: Towards retrieval-augmented large language models. *arXiv preprint*, arXiv:2405.06211. https://arxiv.org/abs/2405.06211

Fuchs, G., Roitman, H., & Mandelbrod, M. (2021). Automatic form filling with Form-BERT. In *Proceedings of the 44th International ACM SIGIR Conference on Research and Development in Information Retrieval (SIGIR '21)*. https://doi.org/10.1145/3404835.3463063

Harries, J.P. (2023). em_german_mistral. GitHub. https://github.com/jphme/EM_German

Izacard, G., Lewis, P., Lomeli, M., Hosseini, L., Petroni, F., Schick, T., Dwivedi-Yu, J., Joulin, A., Riedel, S., & Grave, E. (2022). Atlas: Few-shot learning with retrieval augmented language models. *arXiv preprint*, arXiv:2208.03299. https://arxiv.org/abs/2208.03299

Jeong, C. (2023). Generative AI service implementation using LLM application architecture: Based on RAG model and LangChain framework. *Journal of Intelligent Information Systems*.

Johnson, J., Douze, M., & Jégou, H. (2017). Billion-scale similarity search with GPUs. *arXiv preprint*, arXiv:1702.08734.

LangChain. (2021). Document transformers. Retrieved from

https://python.langchain.com/v0.1/docs/modules/data_connection/document_transformers/

LangChain. (2023). *LangChain documentation*. LangChain. https://langchain.readthedocs.io

LangChain. (2023a). PyPDFLoader documentation. LangChain.

https://api.python.langchain.com/en/latest/document_loaders/langchain_community.document_l
oaders.pdf.PyPDFLoader.html

LangChain. (2023b). CSVLoader documentation. LangChain.

https://api.python.langchain.com/en/latest/document_loaders/langchain_community.document_l
oaders.csv_loader.CSVLoader.html

LangChain. (2023c). RecursiveCharacterTextSplitter documentation. LangChain.

https://api.python.langchain.com/en/latest/character/langchain_text_splitters.character.Recursive
CharacterTextSplitter.html

LangChain. (2023d). CharacterTextSplitter documentation. LangChain.

https://api.python.langchain.com/en/latest/character/langchain_text_splitters.character.Character
TextSplitter.html

LangChain. (2023e). Maximal Marginal Relevance (MMR) documentation. LangChain.

https://api.python.langchain.com/en/latest/example_selectors/langchain_core.example_selectors.
semantic_similarity.MaxMarginalRelevanceExampleSelector.html

Lewis, P., Perez, E., Piktus, A., Petroni, F., Karpukhin, V., Goyal, N., Küttler, H., Lewis, M.,
Yih, W.-t., Rocktäschel, T., Riedel, S., & Kiela, D. (2020). Retrieval-augmented generation for
knowledge-intensive NLP tasks. *arXiv preprint*, arXiv:2005.11401v4.

Liu, Y., Iter, D., Xu, Y., Wang, S., Xu, R., & Zhu, C. (2023). G-Eval: NLG evaluation using GPT-4 with better human alignment. *arXiv preprint*, arXiv:2303.16634.

https://arxiv.org/abs/2303.16634

Nomic AI. (2024). gpt4all. GitHub. https://github.com/nomic-ai/gpt4all

Parsons, K., McCormac, A., Butavicius, M., Pattinson, M., & Jerram, C. (2014). Determining employee awareness using the Human Aspects of Information Security Questionnaire (HAIS-Q). *Computers & Security*, 42, 165-176.

Saenz, M.J., Koufteros, X., Hohenstein, N.O., Feisel, E., Hartmann, E., & Giunipero, L. (2015). Research on the phenomenon of supply chain resilience. *International Journal of Physical Distribution and Logistics Management*, 45(1-2), 90-117.

Streamlit. (2023). *Streamlit: The fastest way to build data apps*. Streamlit. https://streamlit.io

Zhang, T., Kishore, V., Wu, F., Weinberger, K. Q., & Artz, Y. (2020). BERTScore: Evaluating text generation with BERT. *arXiv preprint*, arXiv:1904.09675.